\documentclass[letterpaper]{article} 
\usepackage{aaai23}  
\usepackage{times}  
\usepackage{helvet}  
\usepackage{courier}  
\usepackage[hyphens]{url}  
\usepackage{graphicx} 
\usepackage{natbib}  
\usepackage{caption} 
\usepackage{algorithm}
\usepackage{algorithmic}
\usepackage{multirow}

\usepackage{newfloat}
\usepackage{listings}
\DeclareCaptionStyle{ruled}{labelfont=normalfont,labelsep=colon,strut=off} 
\floatstyle{ruled}
\newfloat{listing}{tb}{lst}{}
\floatname{listing}{Listing}
\title{Don't Be So Sure! Boosting ASR Decoding via Confidence Relaxation}
\author{
    Tomer Wullach,
    Shlomo E. Chazan
}
\affiliations{
    OriginAI, Ramat-Gan, Israel\\

    tomerw, shlomi@originai.co
}

\usepackage{bibentry}

\begin{document}

\maketitle

\begin{abstract}
Automatic Speech Recognition (ASR) systems frequently use a search-based decoding strategy aiming to find the best attainable transcript by considering multiple candidates. One prominent speech recognition decoding heuristic is beam search, which seeks the transcript with the greatest likelihood computed using the predicted distribution. While showing substantial performance gains in various tasks, beam search loses some of its effectiveness when the predicted probabilities are highly confident, i.e., the predicted distribution is massed for a single or very few classes. We show that recently proposed Self-Supervised Learning (SSL)-based ASR models tend to yield exceptionally confident predictions that may hamper beam search from truly considering a diverse set of candidates. We perform a layer analysis to reveal and visualize how predictions evolve, and propose a decoding procedure that improves the performance of fine-tuned ASR models. Our proposed approach does not require further training beyond the original fine-tuning, nor additional model parameters. In fact, we find that our proposed method requires significantly less inference computation than current approaches. We propose aggregating the top $M$ layers, potentially leveraging useful information encoded in intermediate layers, and relaxing model confidence. We demonstrate the effectiveness of our approach by conducting an empirical study on varying amounts of labeled resources and different model sizes, showing consistent improvements in particular when applied to low-resource scenarios.
\end{abstract}

\section{Introduction}

Self-Supervised Learning (SSL) has been widely adopted in a variety of tasks~\cite{devlin2018bert, liu2019roberta, liu2020multilingual, baevski2020wav2vec, liu2020mockingjay} and has shown to be capable of producing representation that can be effectively utilized for downstream tasks. Models trained via SSL leverage a massive amount of unlabeled data during pre-training, and are further tuned for specific tasks using a significantly smaller amount of labeled data. In recent years, SSL has been employed for speech processing tasks~\cite{baevski2020wav2vec, hsu2021hubert, liu2020mockingjay, van2018representation}, utilizing transformer encoders layers~\cite{vaswani2017attention} creating context-aware representations. 

\begin{figure}[ht]
  \includegraphics[width=\linewidth]{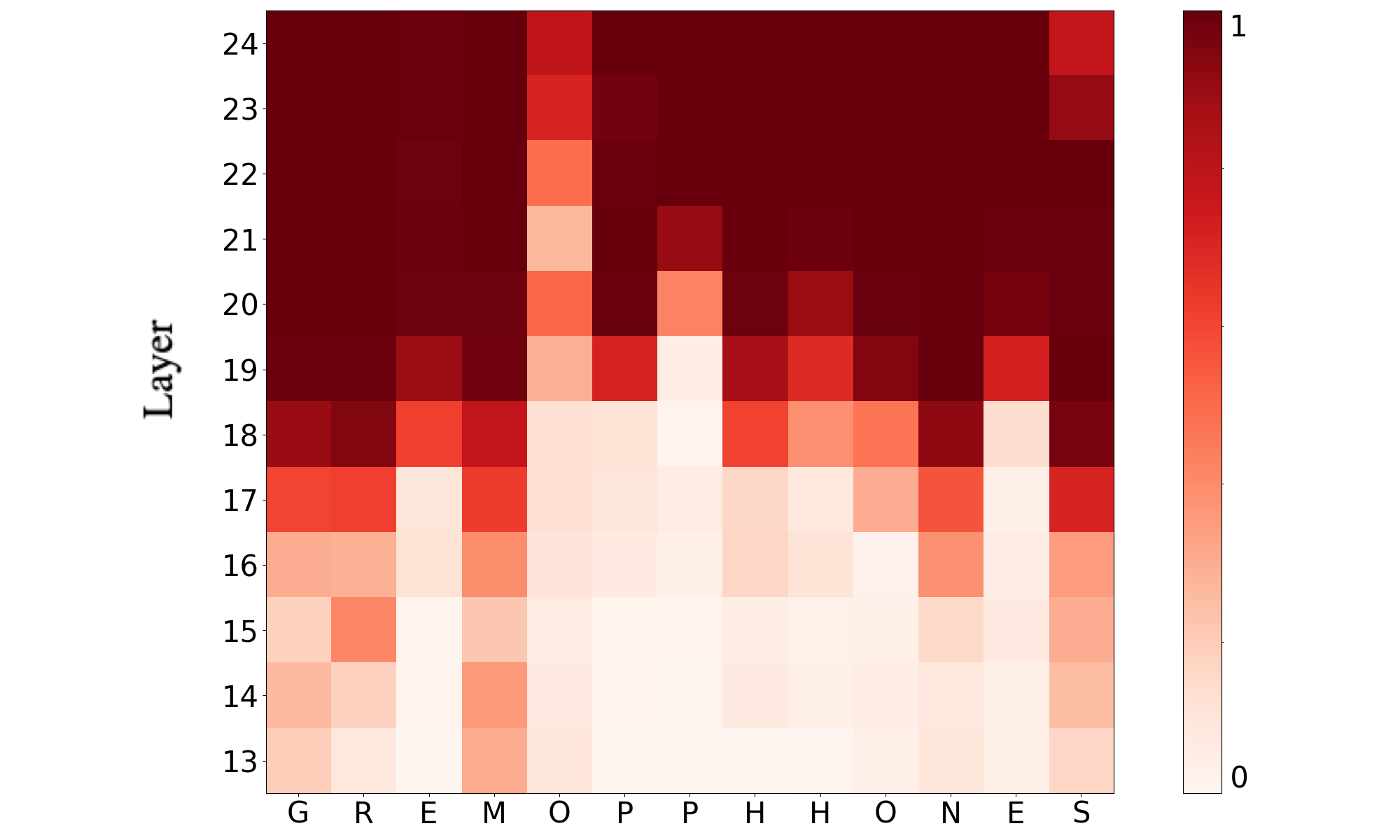}
    \caption{The evolution of confidence throughout the top 12 transformer layers of Wav2vec 2.0 Large fine-tuned on 960 hours of
Librispeech. The confidence increases at the top layers, ultimately reaching exceptionally high levels.}
  \label{fig:evolution}
\end{figure}
 
Recently proposed SSL ASR models commonly employ a greedy decoding strategy, selecting the token with the highest predicted probability at each time step~\cite{baevski2020wav2vec, hsu2021hubert}. This approach is potentially sub-optimal as sequences of high quality might be disregarded. Consequently, researchers often maintain a search-based decoder such as beam search~\cite{baevski2020wav2vec, chen2021wavlm, hsu2021hubert}, showing substantial performance gains, in particular when fused with a Language Model (LM)~\cite{hsu2021hubert, baevski2020wav2vec, chen2021wavlm, gulati2020conformer}.

Beam search considers several alternatives by maintaining the $B$ partial sequences with the highest conditional probability at each decoding step, where $B$ is commonly referred to as Beam Width. Although enabling easy, straightforward decoding, beam search might be prone to biases when predictions are highly confident. In that case, the predicted probability distribution at each time step is massed to very few tokens, thus, reducing the relevancy of the beam width, as a small number of candidates dominate all others. Consequently, highly confident predictions might prevent the algorithm from truly considering multiple alternatives.  

From another perspective, previous works have shown that different layers encode different aspects of the training data~\cite{pasad2021layer, chang2021exploration} such as semantic and acoustic information. Thus, predictions based solely on the top layer might neglect informative features or give excessive weight to others.

\begin{figure*}[t!]
\centering
\includegraphics[scale=0.25]{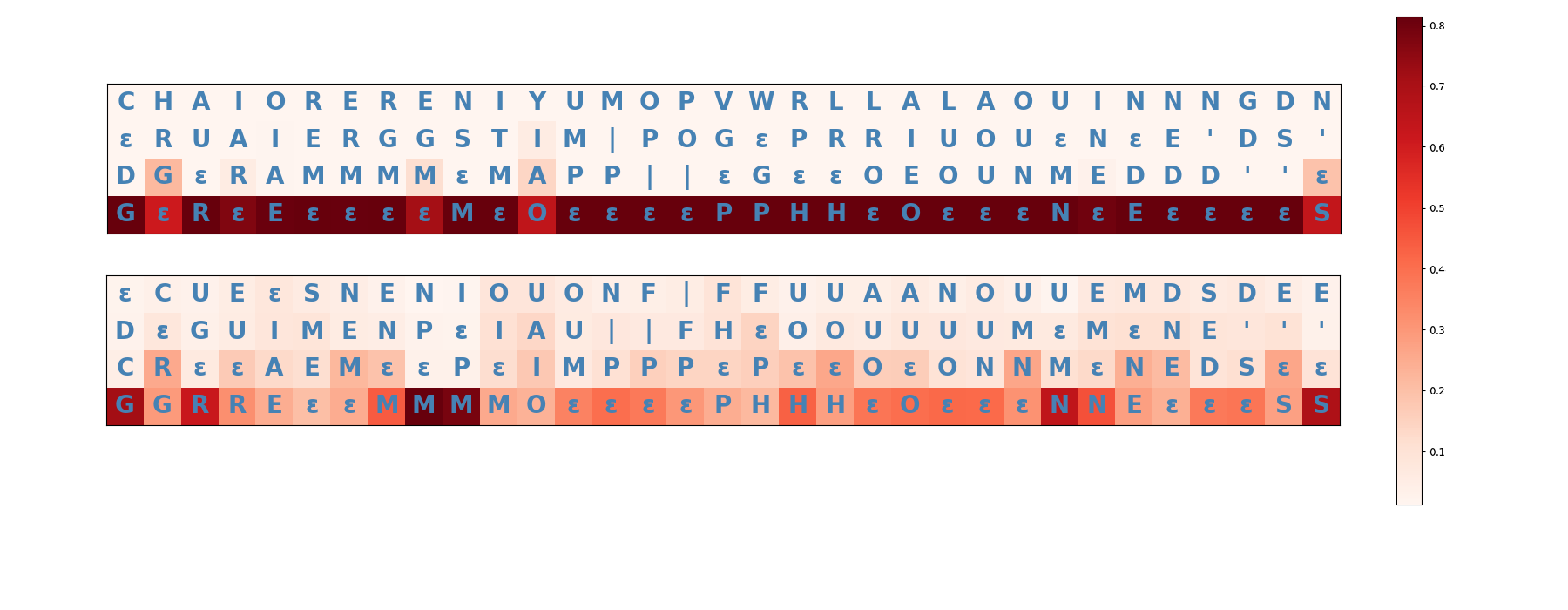}
\caption{\label{fig:logits}Example from Librispeech test-clean: \emph{"...people are not so different from \textbf{gramophones}..."} (wrongly predicted as \emph{gremophones}). Top 4 probabilities for each time step, predicted using wav2vec 2.0 Large fine-tuned using 960 hours of Librispeech data. The model tends to produce exceptionally confident predictions (upper figure). The proposed layer aggregation approach (lower figure) correctly predicted \emph{"gramophones"}. The aggregation relaxed the model's confidence and diversified the decoded candidates.}
\end{figure*}

In this study, we monitor and analyze the confidence levels of several high-performing ASR models, which are composed of a CNN-based encoder followed by Transformer Encoder layers that produce context-aware predictions. In experiments using Wav2vec 2.0~\cite{baevski2020wav2vec} and HuBERT~\cite{hsu2021hubert}, we find that high confidence levels widely emerge at the top transformer layers, while intermediate layers tend to be less confident and more diverse. Furthermore, we evaluate the adverse influence of these findings on the performance of a beam search decoder. 

Recent studies demonstrated the merits of aggregating deep transformer layers~\cite{yang2020novel}, showing that aggregating layers encoding different features create enriched representations that boost performance~\cite{dou2020exploiting, wang2020multi}.
Different from previous works that employed the concept of leveraging intermediate representations~\cite{chang2021exploration, dou2018exploiting,chang2022distilhubert, wang2020multi}, Our method \emph{does not require additional training nor trainable parameters}. Furthermore, we find that the relaxed distribution formed by the aggregation operation enables the use of smaller beam widths without compromising performance, Hence, reducing beam search's computation costs~\cite{meister2020best}. 

Our contributions can be summarized as follows:
\begin{enumerate}
    \item We show that state-of-the-art SSL ASR models produce highly confident predictions, and illustrate how confidence progressively accumulates across layers.
    \item We demonstrate the negative impact of high confidence on speech recognition when using a beam search decoder.
    \item We suggest a layer aggregation approach for relaxing the predicted confidence, harnessing information encoded in intermediate layers, and improving computation costs. We show the benefits of our approach using different amounts of labeled resources and models of sizes.
\end{enumerate}

\section{Related Work}
\paragraph{Self-Supervised Speech Recognition}
Recently proposed SSL speech recognition models follows the large body of work showing tremendous success with employing transformer-based encoders for extracting informative features from raw input. In most cases, these models are first pre-trained using vast amounts of unlabeled data and later fine-tuned using labeled data by solving a supervised training objective. The recently proposed Wav2vec 2.0 model~\cite{baevski2020wav2vec} is composed of a feature encoder that consists of stacked 1-dimensional convolution layers followed by multiple transformer encoder layers. The feature encoder maps raw audio signals into latent speech representations, which are then fed to the first (lowest) transformer encoder layer. The output of each transformer encoder layer is used as input to the next-in-line transformer encoder layer, utilizing the self-attention mechanism to create contextualized speech representations. The model is pre-trained using a contrastive objective, mapping masked spans of a continuous signal into a
discrete set of latent representations using a Gumbel-softmax operation~\cite{jang2016categorical}. The loss is computed by contrasting true quantized latent representations of masked time steps from a set of distractors. Once pre-trained, the model is fine-tuned to speech recognition by projecting each time step's contextualized representation, created by the top transformer encoder layer, into $C$ classes, where $|C|$ is the target vocabulary size. The fine-tuning uses Connectionist Temporal Classification (CTC) loss~\cite{graves2006connectionist} to optimize the model parameters. The CTC loss enables getting around the alignment mismatch of the speech representations and textual labels, as the input audio length is much longer than the target text sequence and most datasets do not provide an accurate alignment of each audio frame and its corresponding text label.
The conditional probability which is used to compute CTC loss is defined as:

\begin{equation}
P_{CTC}(Y \mid X)=\sum_{A \in A_{X,Y}} \prod_{t=1}^{T}{p_{t}(a_t \mid X)},
\label{eq:ctc}
\end{equation}

where $A$ is a set of valid alignments for input sequence $X$ and target sequence $Y$, and $p_{t}(a_t \mid X)$ is the probability of observing label $a_t$ at time $t$.

Different from Wav2vec 2.0 pre-training, Hidden Unit BERT (HuBERT)~\cite{hsu2021hubert} aims to predict the masked speech representation's cluster, assigned using K-means applied to the input's mel-frequency cepstral coefficient features (MFCC). HuBERT uses an offline quantization step to predict, thus, can directly infer target classes rather than relying on a contrastive objective.

\begin{figure}[h]
  \centering
  \includegraphics[scale=0.4]{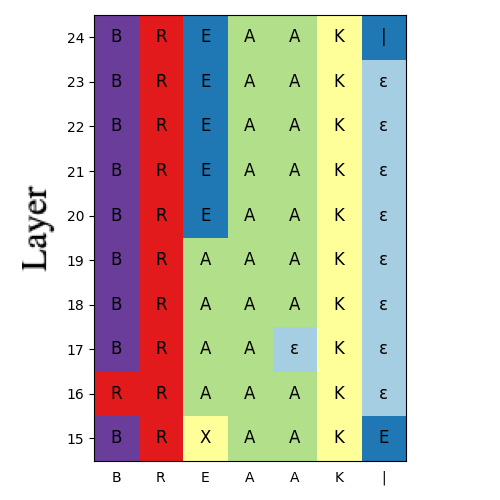}
    \caption{The Prediction evolution across the top 10 transformer layers, using Wav2vec 2.0 Large with a beam search decoder fused with LM. The baseline approach wrongly predicted "break", while the proposed approach was able to produce the correct word "brake".}
  \label{fig:token_evolution}
\end{figure}

\paragraph{Intermediate Transformer Layers Information}
Many studies made efforts to shed light on transformer encoder-based models (BERT-like) mechanisms and the information captured within their intermediate layers~\cite{tenney2019bert, clark2019does}, an area of interest dubbed BERTology~\cite{rogers2020primer}. A key finding is that some features are better represented in certain layers~\cite{shah2021all} and that different layers may encode different input properties. Building on these observations, previous works pointed out the merits of incorporating features encoded in lower layers~\cite{xiong2018multi} and the benefits of fusing multiple layers instead of the traditional approach of using the output produced by the single last layer~\cite{kondratyuk201975,su2020sesamebert}. To summarize the information encoded across layers, \citet{ji2021improving} propose fusing layer representations by sequentially feeding them to a recurrent layer such as LSTM~\cite{hochreiter1997long}, and using its final state as a global representation. In previous work, \citet{wang2020multi} propose using fusion and pooling methods for Neural Machine Translation. They use an aggregation method similar to ours, however, they aggregate layer representations rather than layer logits, and employ the aggregation on the transformer decoder module.

Different from previous works, our method does not require further training nor additional parameters, and applied on models which have  already been pre-trained and fine-tuned. To the best of our knowledge, this is the first work that addresses speech recognition confidence related issues via layer aggregation, and exhibits its effectiveness in both performance and computational costs.

\section{Methodology}
We investigate SSL-based speech recognition models (denoted as Acoustic Models) fine-tuned to predict a transcript sequence of length~$T$, $Y=\left[ y_1,\ldots,y_{T}\right]$, while aiming to minimize a CTC loss~\cite{graves2006connectionist}. The models are fed with an input audio sequence $X$, and output a logit matrix of shape $R^{T \times C}$. The resulting conditional probability can be defined as $P(Y \mid X) \in R^{T \times C}$.

Let $G$ be a speech recognition model, pre-trained using SSL and fine-tuned for speech recognition on dataset $D$. Comprising $G$ are a feature extraction module, that extracts latent representations forming a sequence of speech units $Z$ of length $T$, and a contextual module with $N$ transformer encoder layers. The contextual module is fed with $Z$ and produces contextual representations~\cite{baevski2020wav2vec, hsu2021hubert}. We denote $H^{t}_n$ as the representation produced by transformer layer $n \in 1,\ldots,N$ at time step $t \in 1,\ldots,T$. Let projection\_head($H_N$) be a linear projection layer trained to project the emissions of the top transformer layer, $H_N$, into $C$ classes, where $|C|$ is the vocabulary size. Note that $C$ is composed of alphabet characters as well as special CTC tokens representing blanks and separators denoted as $\epsilon$ and $|$, respectively.


\paragraph{Decoding Strategies}

A greedy decoding heuristic would be observing each position as if it is isolated from the rest of the sequence, and selecting the token that maximizes the probability defined as:

\begin{equation}
  P_{greedy}(Y \mid X) = {\arg\max}_Y \prod_{t = 1}^{T} {p_{t}(y_t \mid X)}.
  \label{eq:argmax}
\end{equation}

However, such paradigm might neglect candidates with higher quality, as previous predictions cannot be altered given newly decoded information.
For this purpose, a beam search decoder is frequently employed, aiming to maximize the combination of both logits and LM scores:

\begin{equation}
    P_{G}(Y \mid X) + \alpha_{1} P_{LM}(Y) + \alpha_{2} |Y|,
    \label{eq:lm}
\end{equation}
where $Y$ is the predicted transcript sequence, $\alpha_1$ denotes the weight of the language model probability and $\alpha_2$ is the word score $|Y|$ is the length of the sequence.

\begin{figure*}[h!]
  \centering
  \includegraphics[scale=0.5]{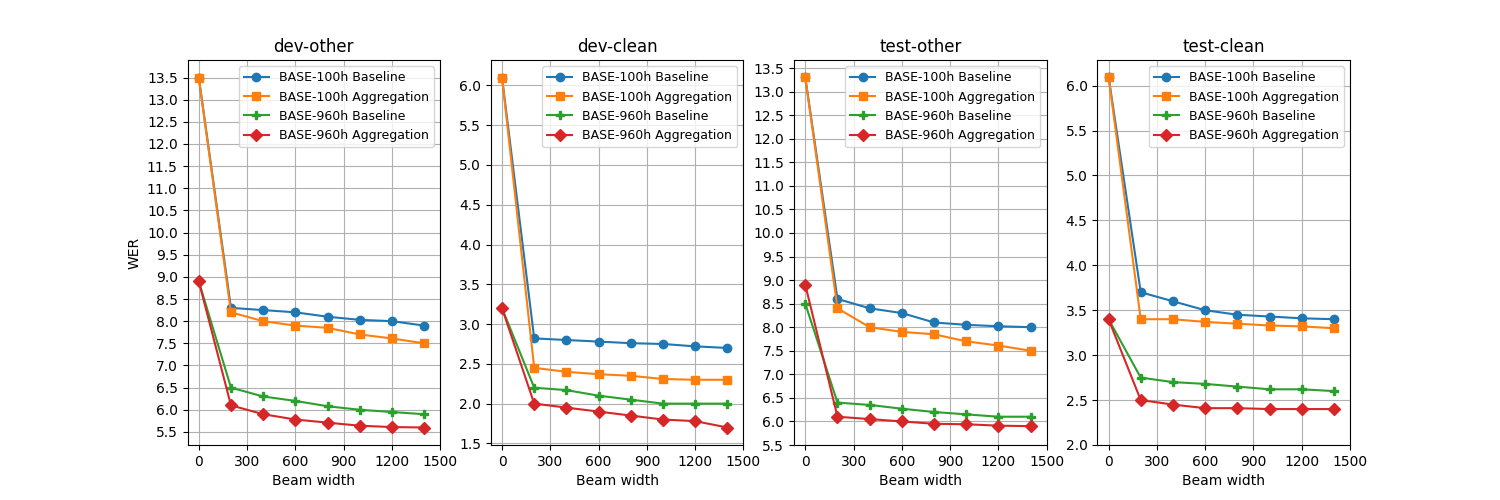}
    \caption{WER as a function of beam width, computed on Librispeech splits using Wav2vec 2.0 BASE fine-tuned using 100h and 960h. The proposed aggregation method is able to perform as well as the standard method, using significantly smaller beam size.}
  \label{fig:beam_width}
\end{figure*}

\paragraph{Confidence Levels}

A network is over-confident when the probability distribution is massed to a single class~\cite{pereyra2017regularizing}. In other words, highly confident predictions occurs when the softmax probabilities computed by employing the projection head on H\_n have low entropy.


To investigate how confidence evolves throughout the transformer layers, we employ the same projection\_head followed by a Softmax operation on the representations emitted by \emph{each transformer layer}.

While projection\_head is trained to project the model's final representations into the vocabulary space, previous works demonstrated that the same projection layer can be employed on intermediate representations as well~\cite{geva-etal-2021-transformer}.

Our findings indicate that there is a confidence increase at the top transformer layer, as illustrated in Figure~\ref{fig:evolution}.
We hypothesize that excessive confidence might prevent beam search from considering promising, yet less confident, predictions and subsequently compromise its performance.

Figure~\ref{fig:logits} (upper) illustrate the top 4 predicted probabilities at each time step. It is worth noting that the predicted probability distribution at the 5-th position is massed almost entirely at token "$E$". 

Completing the picture, the predicted token evolution is illustrated in Figure~\ref{fig:token_evolution}. For each transformer layer at each time step, we chose the token with the largest probability, computed using projection\_head. This illustrates that a correct prediction may live in intermediate layers rather than the top layer. Thus, utilizing intermediate layers can provide beam search with additional valuable information. 


\paragraph{Layer Aggregation.}

Building on these observations, we propose aggregating the logits of the top $M<N$ transformer layers, forming relaxed and better-informed logits to be used by the beam search decoder. The aggregation of the projected logits can be viewed as a projection of the sum of the $M$ layers. Intuitively, this can be accomplished using the distributive property of the matrices, where $A \times B + A \times C = A \times (B+C)$.

As a consequence of the confidence increase across the layers, a scaling mechanism is required to prevent the top layers from dominating the aggregation product. We apply $L2$ normalization on each layer's predicted logits, regularizing the impact of overly confident layers.

Formally, the layer aggregated logits can be described as follows:

\begin{equation}
  \widehat{logits}(X) = \sum_{n=N-M}^{N} projection\_head(\frac{H_n}{||H_n||_2}).
  \label{eq:aggregation}
\end{equation}

\begin{table*}[h]
\centering
  \begin{tabular}{ lccccccccccccccc}
    {} &{} & & {} &\multicolumn{2}{c}{10-min} & \multicolumn{2}{c}{1-hour} & \multicolumn{2}{c}{10-hour} & \multicolumn{2}{c}{100-hour} & \multicolumn{2}{c}{360-hour} & \multicolumn{2}{c}{960-hour}\\\hline
    \multicolumn{2}{c}{Model} & Pre-training& Layers &
    \multicolumn{1}{c}{$\beta$} &  \multicolumn{1}{c}{$M$} & \multicolumn{1}{c}{$\beta$} &  \multicolumn{1}{c}{$M$} & \multicolumn{1}{c}{$\beta$} &  \multicolumn{1}{c}{$M$} & \multicolumn{1}{c}{$\beta$} &  \multicolumn{1}{c}{$M$} & \multicolumn{1}{c}{$\beta$} &  \multicolumn{1}{c}{$M$} & \multicolumn{1}{c}{$\beta$} &  \multicolumn{1}{c}{$M$}
    \\\hline
    \multicolumn{2}{c}{W2v BASE} & LS-960 & 12 & {0.25} & {4} & {0.25} & {4} & {0.3} & {4} & {0.5} & {5} & {0.7} & {4} & {0.75} & {6}\\
    \multicolumn{2}{c}{W2v LARGE} & LS-960 & 24 & {0.5} & {6} & {0.5} & {6} & {0.65} & {6} & {0.7} & {6} & {0.7} & {6} & {0.75} & {12}\\
    \multicolumn{2}{c}{HUBERT BASE} & LS-960 & 12 & {0.5} & {4} & {0.5} & {5} & {0.6} & {5} & {0.65} & {5} & {0.7} & {5} & {0.7} & {6}\\
    \multicolumn{2}{c}{HUBERT LARGE} & LL-60k & 24 & {-} & {-} & {-} & {-} & {-} & {-} & {0.7} & {10} & {0.75} & {12} & {0.75} & {13}\\
    \multicolumn{2}{c}{HUBERT XL} & LL-60k & 48 & {-} & {-} & {-} & {-} & {-} & {-} & {-} & {-} & {-} & {-} & {0.8} & {24}\\
    \hline
  \end{tabular}
  \caption{The Number of aggregated layers ($M$) and aggregation trade-off coefficient ($\beta$) used by the fine-tuned models.}
  \label{tab:params}
\end{table*}

\begin{table*}[h!]
  \begin{tabular}{ llcc|cc|cc|cc|cc|cc|cc|cc}
    \hline
    \multicolumn{2}{l}{} & \multicolumn{4}{c}{Baseline} & \multicolumn{4}{c}{Layer Aggregation} &  \multicolumn{4}{c}{Baseline} & \multicolumn{4}{c}{Layer Aggregation}\\
    \multicolumn{2}{c}{Model} & \multicolumn{2}{c}{test-clean} & \multicolumn{2}{c}{test-other} & \multicolumn{2}{c}{test-clean} & \multicolumn{2}{c}{test-other} &
    \multicolumn{2}{c}{dev-clean} & \multicolumn{2}{c}{dev-other} & \multicolumn{2}{c}{dev-clean} & \multicolumn{2}{c}{dev-other}\\
    \hline
    {} & {} & W & C & W & C & W & C & W & C & W & C & W & C & W & C & W & C\\
    \hline
    \multicolumn{18}{c}{\textbf{10-min labeled}}\\
    \multicolumn{2}{c}{W2v BASE} & 9.1 & 4.2 & 15.6 & 5.9 & 8.3 & 4.0 & 13.5 & 5.9 & 8.9 & 4.5 & 15.7 & 6.0 & 8.1 & 4.0 & 14.2 & 5.7\\
    \multicolumn{2}{c}{W2v LARGE} & 8.9 & 4.1 & 13.1 & 5.5 & 8.1 & 4.0 & 12.2 & 5.2 & 8.6 & 4.4 & 12.9 & 5.4 & 7.9 & 3.9 & 11.7 & 5.1\\
    \multicolumn{2}{c}{HuB BASE} & 9.7 & 4.2 & 15.3 & 5.8 & 8.6 & 4.2 & 13.3 & 5.8 & 9.1 & 4.5 & 15.0 & 5.6 & 8.4 & 4.3 & 7.5 & 3.7\\
    \hline
    \multicolumn{18}{c}{\textbf{1-hour labeled}}\\
    \multicolumn{2}{c}{W2v BASE} & 5.5 & 2.5 & 11.3 & 4.9 & 5.0 & 2.1 & 9.9 & 4.2 & 5.0 & 2.4 & 10.8 & 4.8 & 4.7 & 1.9 & 10.1 & 4.3\\
    \multicolumn{2}{c}{W2v LARGE} & 5.1 & 2.2 & 9.4 & 4.3 & 4.8 & 1.8 & 9.0 & 4.0 & 4.8 & 1.9 & 8.5 & 4.1 & 4.3 & 1.8 & 8.2 & 4.0\\
    \multicolumn{2}{c}{HuB BASE} & 6.1 & 2.8 & 11.3 & 4.8 & 5.7 & 2.4 & 9.8 & 4.3 & 5.6 & 2.3 & 10.9 & 4.8 & 5.1 & 2.4 & 9.2 & 4.0 \\
    \hline
    \multicolumn{18}{c}{\textbf{10-hour labeled}}\\
    \multicolumn{2}{c}{W2v BASE} & 4.3 & 1.8 & 9.5 & 4.1 & 4.0 & 1.7 & 9.0 & 3.9 & 3.8 & 1.5 & 9.1 & 4.0 & 3.7 & 1.5 & 8.9 & 3.8\\
    \multicolumn{2}{c}{W2v LARGE} & 3.8 & 1.5 & 7.3 & 3.5 & 3.6 & 1.6 & 6.9 & 3.2 & 3.4 & 1.4 & 6.9 & 3.3 & 3.2 & 1.3 & 6.8 & 3.2\\
    \multicolumn{2}{c}{HuB BASE} & 4.3 & 1.7 & 9.4 & 4.0 & 4.2 & 1.7 & 9.2 & 3.9 & 3.9 & 1.6 & 9.0 & 4.0 & 3.7 & 1.5 & 3.6 & 1.5 \\
    \hline
    \multicolumn{18}{c}{\textbf{100-hour labeled}}\\
    \multicolumn{2}{c}{W2v BASE} & 3.4 & 1.5 & 8.0 & 3.7 & 3.3 & 1.4 & 7.5 & 3.5 & 2.7 & 1.2 & 7.9 & 3.7 & 2.3 & 1.0 & 7.5 & 3.6\\
    \multicolumn{2}{c}{W2v LARGE} & 2.8 & 1.2 & 6.0 & 2.7 & 2.5 & 1.0 & 5.7 & 2.4 & 2.3 & 1.0 & 5.7 & 2.5 &  2.2 & 0.9 & 5.3 & 2.2\\
    \multicolumn{2}{c}{HuB BASE} & 3.4 & 1.4 & 8.1 & 3.7 & 3.4 & 1.4 & 8.0 & 3.6 & 2.7 & 1.1 & 7.8 & 3.6 & 2.6 & 1.0 & 7.6 & 3.4 \\
    \multicolumn{2}{c}{HuB LARGE} & 3.0 & 1.3 & 7.3 & 3.3 & 2.8 & 1.2 & 6.8 & 2.8 & 2.5& 1.2 & 5.0 & 2.3 & 2.1 & 1.0 & 4.5 & 2.2 \\
    \hline
    \multicolumn{18}{c}{\textbf{360-hour labeled}}\\
    \multicolumn{2}{c}{W2v BASE} & 3.1 & 1.4 & 7.5 & 3.3 & 2.8 & 1.2 & 7.1 & 3.1 & 2.6 & 1.1 & 7.7 & 3.5 & 2.4 & 1.0 & 7.2 & 3.2\\
    \multicolumn{2}{c}{W2v LARGE} & 2.7 & 1.1 & 5.7 & 2.6 & 2.5 & 0.8 & 5.3 & 2.4 & 2.2 & 0.9 & 5.6 & 2.5 & 2.0 & 0.9 & 5.1 & 2.0\\
    \multicolumn{2}{c}{HuB LARGE} & 2.0 & 0.9 & 6.9 & 3.0 & 1.9 & 0.9 & 6.6 & 3.0 & 2.4 & 1.1 & 4.8 & 2.2 & 2.0 & 0.8 & 4.0 &  2.0\\
    \hline
    \multicolumn{18}{c}{\textbf{960-hour labeled}}\\
    \multicolumn{2}{c}{W2v BASE} & 2.6 & 0.76 & 6.1 & 2.5 & 2.4 & 0.75 & 5.9 & 2.4 & 2.0 & 0.7 & 5.9 & 2.8 & 1.7 & 0.5 & 5.6 & 2.3\\
    \multicolumn{2}{c}{W2v LARGE} & 2.3 & 0.64 & 5.0 & 2.0 & 2.1 & 0.63 & 4.9 & 1.9 & 1.7 & 0.2 & 4.6 & 2.0 & 1.5 & 0.2 & 4.3 & 2.0\\
    \multicolumn{2}{c}{HuB BASE} & 2.4 & 0.7 & 6.0 & 2.5 & 2.2 & 0.65 & 5.8 & 2.3 & 1.9 & 0.7 & 5.4 & 2.4 & 1.6 & 0.5 & 5.0 & 2.2\\
    \multicolumn{2}{c}{HuB LARGE} & 1.7 & 0.4 & 3.5 & 1.6 & 1.7 & 0.4 & 3.1 & 1.1 & 1.5 & 0.4 & 3.2 & 1.2 & 1.5 & 0.4 & 3.1 & 1.1\\
    \multicolumn{2}{c}{HuB XL} & 1.6 & 0.4 & 3.0 & 1.1 & 1.4 & 0.4 & 2.9 & 0.9 & 1.7 & 0.4 & 2.6 & 1.0 & 1.4 & 0.3 & 2.3 & 1.0\\
    \hline
  \end{tabular}
  \caption{WER(W) and CER(C) results on Librispeech test and dev sets, computed using the Baseline and the proposed Layer Aggregation approach. The Pre-trained models, Wav2vec 2.0 (W2v) and HuBERT (HuB), were fine-tuned using varying sizes of Librispeech data. The results were obtained using beam search fused with a 4-gram LM.}
  \label{tab:results}
\end{table*}


Finally, the aggregated logits and the top layer's logits are interpolated, forming the logits used for beam search:
\begin{equation}
  \beta \cdot logits(X) + (1-\beta) \cdot \widehat{logits}(X),
  \label{eq:interpolate}
\end{equation}
where $\beta$ is the aggregation trade-off coefficient. 

The following sections provide a comparative study of the aggregation approach effectiveness applied on SSL-ASR models.

\section{Experimental Setup}
Our study includes recently proposed SSL-based models, namely, Wav2vec 2.0 and HuBERT. We experiment with several models variants that differ in size and pre-training data. The model size varies between ~95M and 964M parameters, and pre-training was performed using either 960 hours of Librispeech ('LS-960')~\cite{panayotov2015librispeech} or 60k hours of Libri-Light ('LL-60K')~\cite{kahn2020libri}. 

We conduct our experiments using pre-trained models downloaded from the HuggingFace platform (https://huggingface.co/models), which were fine-tuned using 10m, 1h, 10h, 100h, 360h and 960h 'train' subsets of Librispeech. 

The results for all of the experiments were attained using a beam search decoder with a 4-gram LM provided by pyctcdecode (https://github.com/kensho-technologies/pyctcdecode). The beam search decoder parameters, $\alpha_1$ anf $\alpha_2$ (see eq.~\ref{eq:lm}), were searched and optimized using the Ax toolkit (https://github.com/facebook/Ax). We chose using a 4-gram LM in all of our experiments, which is widely used for speech recognition decoding~\cite{baevski2020wav2vec, hsu2021hubert}, although our proposed aggregation method is not constrained to a specific LM and can be applied using other LM.

We tune the aggregation hyperparameters, namely, the number of aggregated layers ($M$) and aggregation trade-off coefficient ($\beta$) using grid search employed on the 'dev-clean' set. Table~\ref{tab:params} summarizes the aggregation hyperparameters for each of the experimented models.

\paragraph{Speech Recognition Performance.}
We conduct two experimental setups: \emph{Baseline} and \emph{Layer Aggregation}. The baseline setup employs the logits produced using the top layer, while the layer aggregation employs the aggregated logits using the corresponding aggregation parameters. Both setups utilize the same decoder, namely, beam search fused with a 4-gram LMs.

We evaluate both setups using the Librispeech 'test-clean', 'test-other, 'dev-clean', 'dev-other' splits, and report the results using the Word Error Rate(WER) and Charater Error Rate(CER) metrics.

\begin{figure*}[h!]
  \centering
  \includegraphics[scale=0.7]{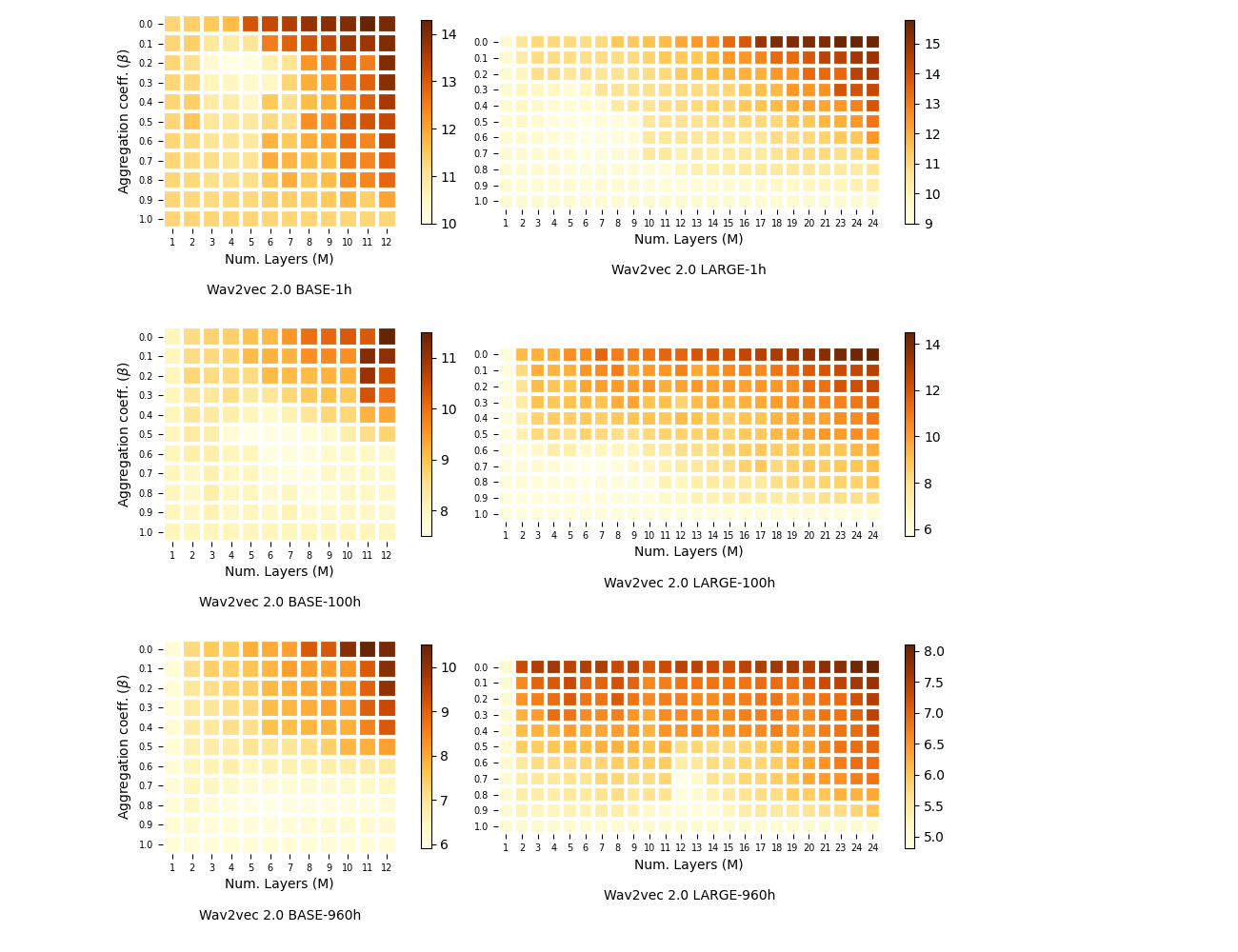}
    \caption{Layer aggregation WER results on Librispeech 'test-other' dataset. The results indicate that low-resourced models (1h and 100h) tend to rely on intermediate layers ($\beta \leq 0.5$), as opposed to models fine-tuned with a large amount of labeled data (960h) that leans towards the top layer.}
  \label{fig:heatmaps}
\end{figure*}


\section{Results}

\paragraph{Speech Recognition Performance on Librispeech.}
Table~\ref{tab:results} reports the results of the Baseline and proposed Layer Aggregation methods.
The results indicate that the proposed approach either matches or improves WER and CER in all experiments. Layer aggregation is particularly effective in lower-resourced settings, i.e. when the amount of labeled data is 10 hours or less. We find that the optimal number of aggregated layers and aggregation trade-off coefficient depends on the size of the model and the amount of labeled data used for fine-tuning, as demonstrated in Table~\ref{tab:params}. It can be observed that models fine-tuned on low-resourced labeled data rely more on the aggregation product rather than the logits produced by the top layer. 

Figure~\ref{fig:heatmaps} exhibits the relationship between the number of aggregated layers ($M$) and the weight given to intermediate layers ($1 - \beta$) by plotting WER values attained using two Wav2vec 2.0 variants, fine-tuned on different amounts of labeled data.
We observed a recurring pattern in the experiments conducted, where models fine-tuned using a small amount of labeled data mostly rely on intermediate layers ($\beta \leq 0.3$), while models fine-tuned using a larger portion of labeled data rely heavily on the top layer's representations.

\paragraph{Computation Efficiency.}
We study the performance of the baseline and the layer aggregation approach using increasing beam width levels, as it is well known that computation costs and beam width are tied~\cite{meister2020best}. To this end, we experiment with two Wav2vec 2.0 BASE variants, fine-tuned using 100 and 960 hours of Librispeech. We set the aggregation parameters, namely, $\beta$ and $M$ according to those detailed in Table~\ref{tab:params}.

Figure~\ref{fig:beam_width} demonstrates the effect of the beam width on both setups.
We observe that while larger beam width typically yields better performance, the aggregation method can match the baseline's WER levels while using a significantly smaller beam width, hence, reducing inference computational costs. For example, when applied to the 'dev-other' set, the aggregation method can match the baseline's performance using a width of 1500, while using a beam width of 400.

\paragraph{Confidence Relaxation.}
To further validate our proposed approach, we compare our  method with another relaxation method, namely, Temperature scaling. For this purpose, we added a temperature scaling step~\cite{guo2017calibration} to the softmax operation of the top layer ($H_N$) and calibrated the temperature parameter using Librispeech 'dev-clean' data. We conducted our experiments using two Wav2vec 2.0 BASE variants, fine-tuned using 100 and 960 hours of Librispeech.
Table~\ref{tab:temperature} shows a comparison of the layer aggregation method with standard Temperature scaling. Note, that besides for a single case, the temperature scaling did not show any significant performance gains. Layer aggregation outperformed temperature scaling in all of the examined cases, suggesting that harnessing intermediate layers is beneficial.

\begin{table}[t]
\begin{tabular}{c|c|c|c}
\hline
Model & \begin{tabular}[c]{@{}c@{}}Fine-tuning\\ Data\end{tabular} & \multicolumn{1}{l|}{Temp.} & \multicolumn{1}{l}{Aggregation} \\ \hline
\multirow{2}{*}{Wav2vec 2.0 BASE}  & 100h                                                       & 7.9                        & 7.5                              \\ \cline{2-4} 
                                   & 960h                                                       & 6.0                        & 5.9                              \\ \hline
\multirow{2}{*}{Wav2vec 2.0 Large} & 100h                                                       & 6.0                        & 5.7                              \\ \cline{2-4} 
& 960h                                                       & 5.0                        & 4.9                              \\ \hline
\end{tabular}
\caption{WER results on Librispeech 'test-other', comparing temperature scaling and the layer aggregation method.}
\label{tab:temperature}

\end{table}

\paragraph{Ablation.}
Examining the errors of models fine-tuned using different sizes of labeled data, we find that the lower-resourced models' errors (fined-tuned using 10m, 1h, 10h) are characterized by spelling errors such as dropping characters or transcribing words exactly the way they sound. The errors of resource-rich models mostly occur with rare words. We noted that the aggregation method successfully assisted with both types of errors. Taking rare words errors for example, using the aggregation method \emph{"Phedrus"} was corrected to \emph{"Phaedrus"}, \emph{"Credius"} to \emph{"Critias"}, and \emph{"Chelsey"} corrected to \emph{"Chelsea"}.

Next, we wish to analyze the choices of the number of aggregated layers and the weight given to intermediate layers.
The trends shown in Figure~\ref{fig:heatmaps} show that models fine-tuned using smaller resources place their aggregation weight on intermediate layers regardless of model size, while the weight is shifted towards the top layers as the amount of fine-tuning labeled data increases. 
These observations are on par with the findings of Pasad \textit{et al.}~\cite{pasad2021layer} which demonstrated that: (a) semantic and acoustic properties are well encoded at the top layers of models fine-tuned using large amounts of labeled data. (b) low-resource fine-tuning is insufficient for diverging the same top layers from the pre-training objective to speech recognition. The latter gives additional motivation to utilize intermediate layers in low-resource setups and further explains why layer aggregation works. 

Furthermore, the proportion of the optimal number of aggregated layers remains the same across model sizes and fine-tuning resources, stretching from 33-41\% of the layers in low-resource models to $\sim$50\% of the layers in resource-rich models.

\section{Consclusions}
In this study, we found that the predictions of SSL-based ASR systems are extremely confident. We then assessed the negative impact of this property on a beam search decoder, while analyzing and visualizing the prediction evolution throughout the transformer layers. Finally, we proposed an aggregation method that aggregates the top transformer layers. Our proposed approach was able to reduce WER and CER with no additional training or parameters, while reducing computational costs.

\bibliography{aaai23.bib}

\end{document}